\documentclass[sigconf]{acmart}

\renewcommand\footnotetextcopyrightpermission[1]{} 
\AtBeginDocument{%
  }

\setcopyright{none}
\copyrightyear{}
\acmYear{2024}
\acmDOI{}

\acmConference[SC'24 TPC Workshop]{Accelerating the Development and Use of Generative AI for Science and Engineering: The Trillion Parameter Consortium (TPC)}{November 2024}{Atlanta, GA, USA}

\begin{document}

\title{Balancing Speed and Stability: The Trade-offs of FP8 vs. BF16 Training in LLMs}

\author{Kazuki Fujii}
\email{kazuki.fujii@rio.gsic.titech.ac.jp}
\authornotemark[1]
\affiliation{%
  \institution{Tokyo Institute of Technology}
  \city{Tokyo}
  \country{Japan}
}

\author{Taishi Nakamura}
\email{taishi.nakamura@rio.gsic.titech.ac.jp}
\affiliation{%
  \institution{Tokyo Institute of Technology}
  \city{Tokyo}
  \country{Japan}
}

\author{Rio Yokota}
\email{rioyokota@rio.gsic.titech.ac.jp}
\affiliation{%
  \institution{Tokyo Institute of Technology}
  \city{Tokyo}
  \country{Japan}
}

\begin{abstract}
Large Language Models (LLMs) have attracted significant attention due to their human-like language understanding and generation capabilities, as well as their applicability across various domains.

These models, characterized by their massive scale and extensive training data, continue to push the boundaries of what is possible in natural language processing. The Llama 3 series, for instance, exemplifies this trend with its flagship model boasting 405 billion parameters trained on 15.6 trillion tokens~\cite{dubey2024llama3herdmodels}.
The immense computational demands associated with training such models have spurred ongoing research into optimizing the efficiency of the training process, particularly through the use of lower-precision formats.

NVIDIA's H100 GPU, which introduces support for FP8 in addition to the more conventional FP16 and BF16 formats, has emerged as a focal point in this optimization effort. Preliminary studies suggest that FP8 could offer substantial reductions in training time without sacrificing model performance when compared to BF16~\cite{micikevicius2022fp8formatsdeeplearning}, making it a promising candidate for large-scale model training. However, the broader implications of adopting FP8, particularly in terms of training stability and downstream task performance, have yet to be fully understood.

In this study, we delve into the practical trade-offs involved in adopting FP8 over BF16 for training LLMs.
We used Megatron-LM\footnote{\url{https://github.com/NVIDIA/Megatron-LM}} to conduct continued pre-training of Llama-3-70B on multilingual and code corpus of approximately 100 billion tokens.
We compared BF16 and FP8 training in terms of training throughput (measured in TFLOPS), training loss behavior, and performance across a range of downstream tasks in both Japanese and English. Our results show that FP8 training improved the training speed from 415 TFLOPS (with BF16) to a maximum of 570 TFLOPS. However, the FP8 settings shown in Table~\ref{tab:fp8_settings} led to unstable training loss and frequent loss spikes, as illustrated in Figure~\ref{fig:fp8-bf16}.

Moreover, the impact of FP8 on downstream task performance was found to be uneven across different task categories. In Japanese, for instance, tasks such as question answering (QA) appeared relatively resilient to the precision format change, with minimal performance degradation observed. In contrast, tasks requiring code generation and mathematical reasoning suffered more pronounced declines, highlighting the sensitivity of certain task types to the precision format used during training. Similar patterns were observed in English tasks, though the distinction between task categories was less stark.

\begin{table}[h]
\centering
\caption{FP8 configuration settings}
\small
\begin{tabular}{|l|l|}
\hline
\textbf{Parameter}              & \textbf{Value}    \\ \hline
\texttt{fp8}                    & \texttt{hybrid}   \\ \hline
\texttt{fp8\_margin}            & \texttt{0}        \\ \hline
\texttt{fp8\_interval}          & \texttt{1}        \\ \hline
\texttt{fp8\_amax\_history\_len} & \texttt{1}        \\ \hline
\texttt{fp8\_amax\_compute\_algo} & \texttt{most\_recent} \\ \hline
\texttt{fp8\_wgrad}             & \texttt{True}     \\ \hline
\end{tabular}
\label{tab:fp8_settings}
\end{table}
\end{abstract}

\maketitle

\begin{figure}[ht]
    \centering
    \includegraphics[width=1.0\linewidth]{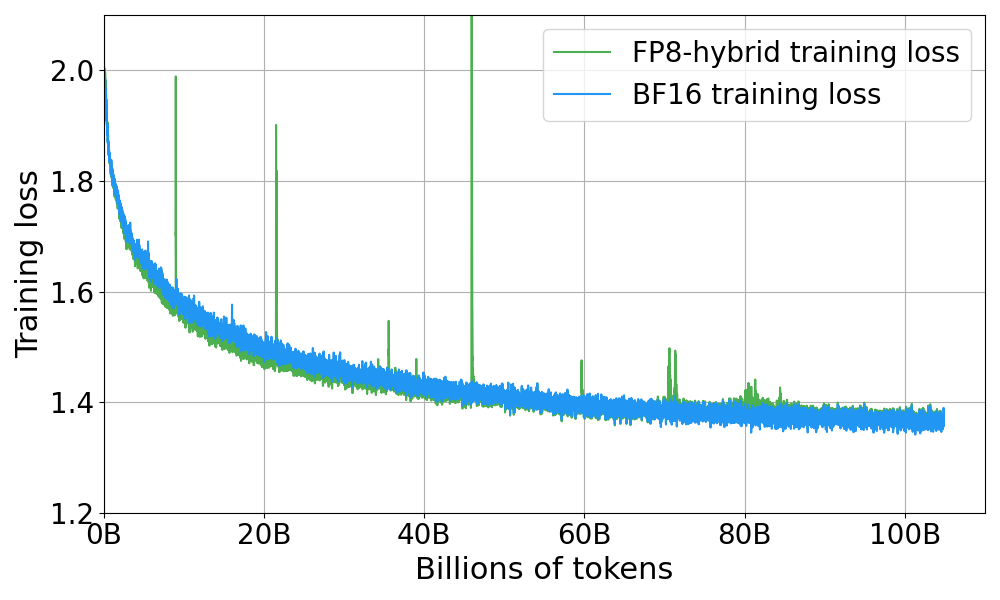}
    \caption{Training loss curves}
    \label{fig:fp8-bf16}
    \Description{}
\end{figure}

\begin{figure}[ht]
    \centering
    \includegraphics[width=1.0\linewidth]{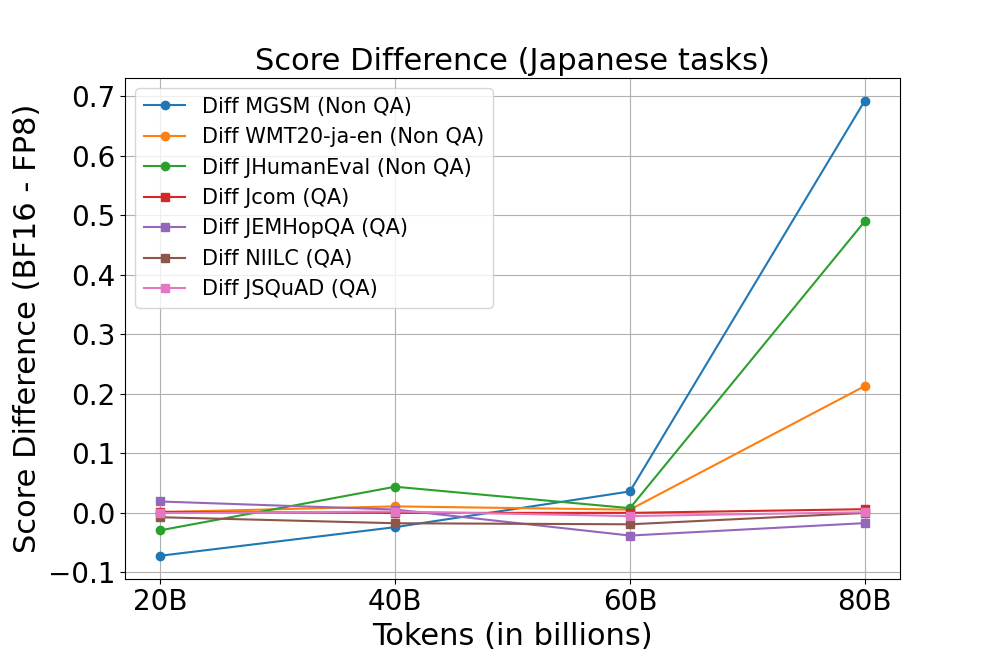}
    \caption{FP8-BF16 score difference in Japanese tasks}
    \label{fig:fp8-bf16-eval-diff-ja}
    \Description{}
\end{figure}

\bibliographystyle{plain}
\bibliography{sample-base}

\appendix

\section{Reproducibility statement}

To ensure the reproducibility of our experiments, we provide detailed information about the software versions used in this study and training settings.

\begin{itemize}
    \item We conducted our experiments using Megatron-LM with Megatron-core version 0.6.0.
    \item The TransformerEngine\footnote{\url{https://github.com/NVIDIA/TransformerEngine}} version used in our study is v1.6
\end{itemize}

By providing this information, we aim to facilitate the replication of our experiments and encourage further research in this area.

\subsection{Training settings}

The hyperparameters shown in Table \ref{tab:hyperparameters} were utilized for both BF16 and FP8 experiments.
The AdamW optimizer was employed for training the models, with hyperparameters $\beta_1=0.9$, $\beta_2=0.95$, $\epsilon=1.0 \times 10^{-8}$.
A cosine learning rate scheduler was used, and the learning rate was set to reach its maximum value at 1,000 warmup steps and finally decay to 1/10 of that value.
Furthermore, Flash Attention 2 was adopted for improved computational efficiency and memory footprints.

\begin{table}[h]
\centering
\caption{Training hyperparameters}
\label{tab:hyperparameters}
\begin{tabular}{|l|c|}
\hline
\textbf{Hyperparameter} & \textbf{Value} \\
\hline
Learning Rate (LR) & 1.0E-5 \\
Minimum Learning Rate & 1.0E-6 \\
Global Batch Size & 1024 \\
Weight Decay & 0.1 \\
Gradient Clipping & 1.0 \\
\hline
\end{tabular}
\end{table}

\subsection{Training environment}

We conducted training experiments using BF16 and FP8 precision on different high-performance computing infrastructures. For the BF16 experiments, we utilized the AI Bridging Cloud Infrastructure (ABCI) of the National Institute of Advanced Industrial Science and Technology, Japan. We used multiple NVIDIA A100 nodes, each equipped with eight NVIDIA A100 40GB GPUs, interconnected via InfiniBand HDR.
For the FP8 experiments, we utilized the TSUBAME 4.0 supercomputer at the Tokyo Institute of Technology Global Scientific Information and Computing Center. In this setting, we employed NVIDIA H100 SXM5 94GB GPUs, with each node containing four H100 GPUs, interconnected via InfiniBand NDR200.

\subsection{Distributed training}

\begin{table}[ht]
    \begin{center}
    \caption{Distributed training settings. DP, TP, PP, and SP represent Data Parallelism, Tensor Parallelism, Pipeline Parallelism, and Sequence Parallelism, respectively.}
    
    \begin{tabular}{rrrrcc}
        \toprule
        Model & DP & TP & PP & SP & Distributed Optimizer \\ \midrule
        70B (BF16) & 2 & 8 & 16 & \checkmark & \checkmark \\
        70B (FP8) & 4 & 4 & 8 & \checkmark & \checkmark \\
        \bottomrule
    \end{tabular}
    \label{tab:distributed_settings}

    \end{center}
\end{table}

To efficiently perform the training process, we adopted 3D parallelism, which integrates data parallelism, tensor parallelism, and pipeline parallelism, aiming for high computational efficiency and efficient memory utilization.
We used the Megatron-LM\footnote{\url{https://github.com/NVIDIA/Megatron-LM}} library for training.
Table \ref{tab:distributed_settings} shows the distributed training settings.

\section{Downstream evaluation result}

\begin{figure}[ht]
    \centering
    \includegraphics[width=1.0\linewidth]{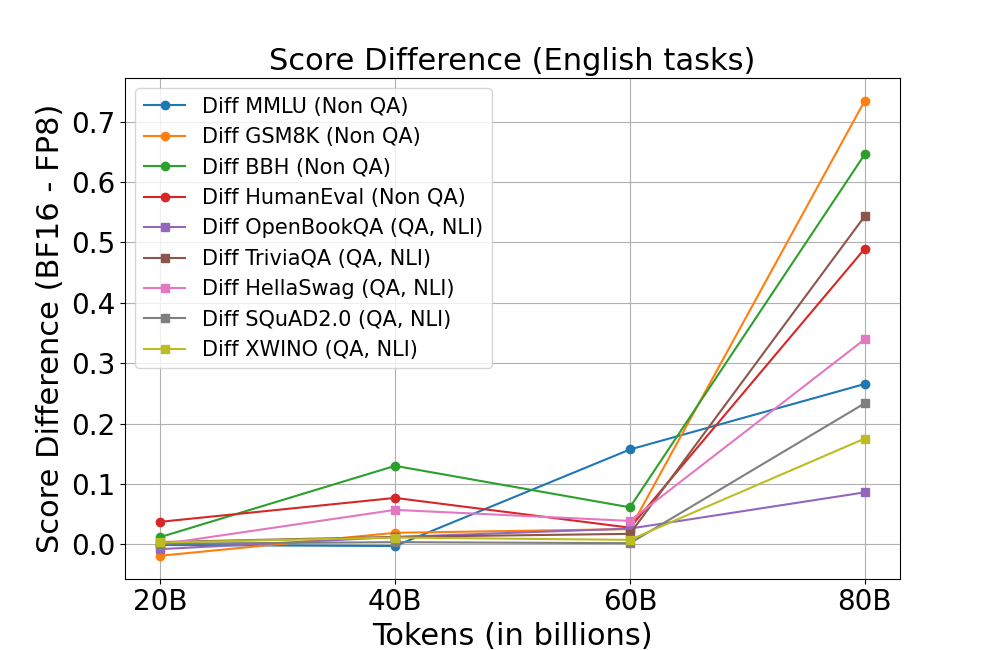}
    \caption{FP8-BF16 Score difference in English tasks}
    \label{fig:fp8-bf16-eval-diff-en}
    \Description{}
\end{figure}

The results of introducing FP8 in Japanese tasks were shown in Figure~\ref{fig:fp8-bf16-eval-diff-ja}, and here, we present the corresponding results for English tasks in Figure~\ref{fig:fp8-bf16-eval-diff-en}.

\section{Evaluation tasks}

The evaluation methods for Japanese are shown in Tables~\ref{tab:eval_benchmark_ja}. 
The dataset consists of six types of Japanese tasks, with few-shot settings for question answering (QA), reading comprehension (RC), arithmetic reasoning (AR), machine translation (MT), general knowledge, and code generation.

\begin{table}[h]
    \begin{center}
    \footnotesize
    \setlength{\tabcolsep}{1pt}

    \begin{tabular}{l|cccccccc}
        \toprule
        Benchmark & \multicolumn{4}{c}{\texttt{llm-jp-eval} (v1.3.0)} & \multicolumn{3}{c}{\texttt{JP LM Evaluation Harness}\footnotemark{}} \\
        Eval. task  & \multicolumn{3}{c}{Question Answering} & \multicolumn{1}{c}{RC} & \multicolumn{1}{c}{AR} & \multicolumn{1}{c}{Machine Translation} \\
        Dataset & \multicolumn{1}{c}{Jcom} & \multicolumn{1}{c}{JEMHQA} & \multicolumn{1}{c}{NIILC} & \multicolumn{1}{c}{JSQuAD} & \multicolumn{1}{c}{MGSM}  & \multicolumn{1}{c}{WMT'20\textsubscript{Ja-En}} \\ \midrule
        Instances & 1,119 & 120 & 198 & 4,442 & 250 & 993 \\
        Few-shots & 4 & 4 & 4 & 4 & 4 & 4 \\
        Eval. metric & EM acc. & Char-F1 & Char-F1 & Char-F1 & EM acc. & \multicolumn{1}{c}{BLEU} \\
        \bottomrule
    \end{tabular}
    \vspace{10pt}
    
    \begin{tabular}{l|cc}
        \toprule
        Benchmark & \texttt{llm-jp-eval} (v1.3.0) &  \\
        Eval. task & General Knowledge & Code Generation \\ 
        Dataset & JMMLU & JHumanEval \\ \midrule
        Instances & 7,097 & 164 \\
        Few-shots & 5 & 0 \\
        Eval metrics & Accuracy & pass@1 (n=10) \\
        \bottomrule
    \end{tabular}

    \caption{\label{tab:eval_benchmark_ja}Japanese datasets and additional benchmarks}
    \end{center}
\end{table}

\footnotetext{\url{https://github.com/Stability-AI/lm-evaluation-harness}}
\end{document}